\documentclass{mic2015}

\begin{document}

\title{An intelligent extension of Variable Neighbourhood Search for labelling graph problems
}
\author{Sergio Consoli\inst{1} \and Jos\'{e} Andr\'{e}s Moreno P\'{e}rez\inst{2}}
\institute{
 ISTC/STLab, National Research Council (CNR), Catania, Italy\\
  \email{sergio.consoli@istc.cnr.it}
  \and
  Department of Computing Engineering, Universidad de La Laguna, Tenerife, Spain\\
  \email{jamoreno@ull.edu.es}
}
\id{id}
\maketitle

\begin{abstract}
In this paper we describe an extension of the Variable Neighbourhood Search (VNS) which integrates the basic VNS
with other complementary approaches from machine learning, statistics and experimental algorithmic,
in order to produce high-quality performance and to completely automate the resulting optimization strategy.
The resulting intelligent VNS has been successfully applied to a couple of optimization problems where the solution space consists of the subsets of a finite reference set.
These problems are the labelled spanning tree and forest problems that are formulated on an undirected labelled graph; a graph where each edge has a label in a finite set of labels $L$.
The problems consist on selecting the subset of labels such that the subgraph generated by these labels has an optimal spanning tree or forest, respectively.
These problems have several applications in the real-world, 
where one aims to ensure connectivity by means of homogeneous connections.
\end{abstract}

\section{Introduction}

In this paper we scratch an Intelligent Variable Neighbourhood
Search (Int-VNS) aimed to achieve further improvements of a
successful VNS for the Minimum Labelling Spanning Tree (MLST) and
the $k$-Labelled Spanning Forest ($k$LSF) problems. This approach
integrates the basic VNS with other complementary intelligence tools
and has been shown a promising strategy in~\cite{MLSTP-intell-ASOC}
for the MLST problem and in~\cite{MLSTP-int-INOC} for the $k$LSF
problem. The approach could be easily adapted to other optimization
problems where the space solution consists of the subsets of a
reference set; like the feature subset selection or some location
problems. First we introduced a local search mechanism that is
inserted at top of the basic VNS to get the Complementary Variable
Neighbourhood Search (Co-VNS). Then we insert a probability-based
constructive method and a reactive setting of the size of shaking
process.

\section{The Labelled Spanning Tree and Forest problems}
A labelled graph $G=(V,E,L)$ consists of an undirected graph where
$V$ is its set of nodes and $E$ is the set of edges that are labelled on the set $L$ of labels.
In this paper we consider two problems defined on a labelled graph:
the MLST and the $k$LSF problems. The MLST problem~\cite{Chang}
consists on, given a labelled input graph $G=(V,E,L)$, to get the
spanning tree with the minimum number of labels; i.e., to find the
labelled spanning tree $T^*=(V,E^*,L^*)$ of the input graph that
minimizes the size of label set $|L^*|$. The $k$LSF
problem~\cite{CerulliKLSF} is defined as follows. Given a labelled
input graph $G=(V,E,L)$ and an integer positive value $\bar{k}$, to
find a labelled spanning forest $F^*=(V,E^*,L^*)$ of the input graph
having the minimum number of connected components with the upper
bound $\bar{k}$ for the number of labels to use, i.e. $min
|Comp(G^*)|$ with $|L^*| \leq \bar{k}$. Given the subset of labels
$L^* \subseteq L$, the labelled subgraph $G^*=(V,E^*,L^*)$ may
contain cycles, but they can arbitrarily break each of them by
eliminating edges in polynomial time until a forest or a tree is
obtained. Therefore in both problems, the matter is to find the
optimal set of labels $L^*$.
Since a MLST solution 
would be a solution also to the $k$LSF problem if the obtained solution tree would not violate the limit $\bar{k}$ on the used number of labels, it is easily deductable that the two problems are deeply correlated.
The NP-hardness of the MLST and $k$LSF problems was stated in
\cite{Chang} and in~\cite{CerulliKLSF} respectively. Therefore any
practical solution approach to both problems requires
heuristics~\cite{CerulliKLSF,VNSkLSF2014}.

\section{Complementary Variable Neighbourhood Search}
\label{sub_sec_ComplLS}

The first extension of the VNS metaheuristic that we introduced for these problems is a 
local search mechanism that is inserted at top of the basic
VNS~\cite{VNSkLSF2014}. The resulting local search method is
referred to as~\emph{Complementary Variable Neighbourhood Search}
(Co-VNS)~\cite{MLSTP-int-INOC,MLSTP-intell-ASOC}.
Given a labelled graph $G =
(V,E,L)$ with $n$ vertices, $m$ edges, and $\ell$ labels, Co-VNS
replaces iteratively each incumbent solution $L^*$
with another 
solution selected from the \emph{complementary space} of $L^*$ defined as the sets of labels that are not contained in $L^*$; $L \setminus L^*$. 
The iterative process of extraction 
of a complementary solution
helps to escape the algorithm from possible traps 
in local minima, since the complementary solution lies 
in a very different zone of the search space with respect to the
incumbent solution. This process yields 
an immediate peak of diversification 
of the whole local search procedure. 
To get a complementary 
solution, Co-VNS uses a greedy heuristic as
constructive method in the complementary space of the current solution.
For the MLST and $k$LSF problems the greedy heuristic is the Maximum Vertex Covering Algorithm (MVCA)~\cite{Chang}
applied to the subgraph of $G$ with labels in
$L\setminus L^*$.
Note that Co-VNS stops if either
the set of unused labels
contained in the complementary space is empty ($L\setminus L^*=
\emptyset$) or a final feasible solution is produced.
Successively, the basic VNS 
is applied in order to improve the resulting solution.

At the starting point of VNS, it is required to define a suitable
series of neighbourhood structures of size $q_{max}$. 
In order to impose a
neighbourhood structure on the solution space $S$
we use the Hamming distance between two
solutions $L_1, L_2 \in S$ given by
$\rho(L_1,L_2)=|L_1 \Delta L_2|$
where $L_1 \Delta L_2$
consists of labels that are in one of the
solutions but not in the other.
VNS starts from an initial solution $L^*$ with $q$
increasing iteratively from 1 up to the maximum neighborhood size, $q_{max}$. 
The basic idea of VNS to change the neighbourhood structure when the
search is trapped at a local minimum, is implemented by the shaking phase.
It consists of the random selection of another point in the neighbourhood $N_q(L^*)$ of the current solution $L^*$.
Given $L^*$, we consider its $q^{th}$ neighbourhood $N_q(L^*)$ comprised by
sets having a Hamming distance from $L^*$ equal to $q$ labels, where
$q = 1, 2,\ldots, q_{max}$.
In order to construct the neighbourhood of a solution $L^*$, the
algorithm proceeds with the deletion of $q$ labels from $L^*$.
\section{Intelligent Variable Neighbourhood Search}
\label{sub_sec_hybrid}

The proposed intelligent metaheuristic (Int-VNS) is built 
from Co-VNS, with the
insertion of a probability-based local search as constructive method
to get the complementary space solutions.
In particular, this local search is a modification of greedy heuristic,
obtained by introducing a probabilistic choice on the next label to be added into incomplete solutions.
By
allowing worse components to be added to incomplete solutions, this
probabilistic constructive heuristic produces a further increase on the diversification of the optimization process. The construction
criterion is as follows. The procedure starts from an initial
solution and iteratively selects at random a candidate move. If this
move leads to a solution having a better objective function value
than the current solution, then this move is accepted
unconditionally; otherwise the move is accepted with a probability
that depends on the deterioration, $\Delta$, of the objective
function value.
This construction criterion takes inspiration from Simulated
Annealing (SA)~\cite{NonMonotonicSA}. 
However, the
probabilistic local search works with partial solutions which are
iteratively extended with additional components until complete
solutions emerge. 
In the probabilistic local search, the acceptance probability of a
worse component into a partial solution is evaluated according to
the usual SA criterion by the Boltzmann function $\exp(-\Delta /T)$,
where the temperature parameter $T$ controls
the dynamics of the search. Initially the value of $T$ is large,
so allowing many worse moves to be accepted, and is gradually
reduced by the following geometric cooling law:
$T_{j+1} = \alpha \cdot T_j$, where $T_{0}=|Best_L|$ and $\alpha=1/|Best_L| \in [0, 1]$, with $Best_L$ being the current best solution.
This cooling law is very fast and produces a good balance between
intensification and diversification capabilities. In addition, this
cooling schedule does not requires any intervention from the user
regarding the setting of its parameters, as it is guided
automatically by the best solution $Best_L$. Therefore the whole
optimization process is able to react in response to the search
algorithm's behavior and to adapt its setting on-line according to
the instance of the problem under evaluation~\cite{NonMonotonicSA}.
The probabilistic local search has the purpose of allowing also the
inclusion 
of less promising labels to incomplete solutions.
Probability values
assigned to each label are decreasing in the quality of the solution they give.
In this way, at each step,
labels with a better quality will have a higher
probability of being selected;
the progressive reduction of the temperature 
in the adaptive cooling law produces, step by step, an increasing of
this diversity 
in probabilities.

At the beginning of Int-VNS, the algorithm generates an initial
feasible solution at random, that is the first current best solution $Best_L$, and set parameter
$q_{max}$ to the number of labels of the initial solution ($q_{max}
\leftarrow |Best_L|$).
Then the \emph{Complementary} procedure is applied to $Best_L$ to obtain a solution $L^*$ from the
complementary space of $Best_L$ by means of the probabilistic local search.
The Complementary procedure stops if either a feasible solution $L^*$
is obtained, or the set of unused labels contained in the
complementary space is empty
producing a final infeasible solution.
Subsequently, the 
shaking phase used for the basic VNS
is applied to the resulting
solution $L^*$. 
It consists of the random
selection of a point $L'$ in the neighbourhood $N_q(L^*)$ of the
current solution $L^*$, as in Co-VNS. 
The successive local search corresponds also to that of Co-VNS. 
However, since 
either Co-VNS, or the
deletion of labels in the shaking phase, can produce an 
incomplete solution, the first step of the local search consists of
including additional labels in the current solution in order to
restore feasibility, if needed. The addition of labels at this step
is according to the probabilistic procedure. 
Then, the local search tries to drop labels in $L'$, and then to add further labels following the greedy rule,
until $\bar{k}$ labels emerge.
At this step, if no improvements are obtained the
neighbourhood structure is increased ($q \leftarrow q + 1$)
producing progressively a larger diversification.
Otherwise, the algorithm moves $L^*$ to solution $L'$ restarting the search
with the smallest neighbourhood ($q \leftarrow 1$). 
This iterative process is repeated until the maximum size of the
shaking phase, $q_{max}$, is reached. The resulting local minimum
$L^*$ is compared to the current best solution $Best_L$, which is
updated in case of improvement ($Best_L\leftarrow L^*$).
At this point a reactive setting for the parameter $q_{max}$ is used~\cite{ReactiveVNS}. 
In case of an improved solution, $q_{max}$
is decreased ($q_{max} \leftarrow max(q_{max}-1; |Best_L|/2 )$) in
order to raise the intensification factor of the search process.
Conversely, in case of none improvement, the maximum size of the
shaking is increased ($q_{max} \leftarrow min(q_{max}+1; 2 \cdot
|Best_L|))$ in order to enlarge the diversification factor of the
algorithm. In each case, the adaptive setting of $q_{max}$ is
bounded to lie in the interval between $|Best_L|/2$ and $2 \cdot |Best_L|$ 
to avoid a lack of balance between intensification and
diversification factors.
The algorithm proceeds with the same procedure until the user
termination conditions 
are satisfied, producing at the end the best solution to date, $Best_L$, as
output.

\section{Summary and Outlook}

The achieved optimization strategy seems to be highly promising for both labelling graph problems.
Ongoing investigation consists in statistical comparisons of the
resulting strategy against the best algorithms in the
literature for these problems, in order to quantify and qualify the improvements
obtained.
Further investigation will deal with the application of this strategy to other problems.


\begin{scriptsize}

\end{scriptsize}

\end{document}